%% file: main.tex
\documentclass[11pt]{article}

\usepackage[a4paper,margin=1in]{geometry}
\usepackage{amsmath,amssymb}
\usepackage{booktabs}
\usepackage{array}
\usepackage{longtable}
\usepackage{multirow}
\usepackage{graphicx}
\usepackage{tikz}
\usepackage{float}
\usepackage{natbib}
\usepackage[hidelinks]{hyperref}
\usepackage{enumitem}

\usetikzlibrary{arrows.meta,positioning,shapes.geometric,calc,decorations.pathreplacing}
\graphicspath{{figures/}}
\numberwithin{figure}{section}
\numberwithin{table}{section}

\newcommand{\benchmark}{CruiseBench}
\newcommand{\datasetname}{CPM-N-CMAPSS}
\newcommand{\mainsetting}{CruiseBench-eta5-W256-S10}

\title{\benchmark: A Real-Flight-Aligned N-CMAPSS Benchmark for Engine RUL Prediction}
\author{Pu Cheng\\\href{mailto:cheng.chengpu123@gmail.com}{cheng.chengpu123@gmail.com}
\and
Qiang Miao\\\href{mailto:mqiang@scu.edu.cn}{mqiang@scu.edu.cn}}
\date{\today}

\begin{document}

\maketitle

\begin{abstract}
Remaining useful life (RUL) prediction estimates how long an engine can continue safe operation and is central to maintenance planning.
N-CMAPSS extends C-MAPSS by simulating run-to-failure aero-engine trajectories using recorded real-flight profiles and retaining complete within-flight time series rather than cycle-level snapshots. However, this added realism reduces evaluation control because full-flight records increase data volume and entangle degradation cues with operating-regime variation, complicating preprocessing choices and direct comparisons of RUL modeling performance.
To mitigate this issue, this paper proposes \benchmark, a cruise-stage RUL benchmark derived from N-CMAPSS.
It introduces \datasetname\ (Cruising-Period Mask for N-CMAPSS), a mask artifact that stores cycle-local cruising intervals identified by the common-altitude method for the nine accessible subdatasets. \benchmark\ applies a fixed protocol to the masked rows, using scenario descriptors and measured sensors as inputs while excluding virtual sensors, health parameters, and auxiliary metadata from the feature tensor, preserving native-resolution windows, and applying dataset-wise RUL caps. Experiments with LSTM, GRU, TCN, and TSMixer provide baseline results for this setting. Under \mainsetting, TSMixer obtains the lowest average RMSE, $3.46\pm1.71$, and Saxena score, $(2.50\pm2.99)\times10^4$. Ablation studies show that flight-stage selection, temporal downscaling method, and RUL-cap threshold affect reported results. With its fixed cruise-stage protocol, \benchmark\ provides a reproducible sub-benchmark for controlled RUL model comparison and \datasetname\ provides a stage-specific data foundation for future transfer-learning and domain-adaptation studies. The code is available at \url{https://github.com/NostalgiaJohn/CruiseBench}.

\end{abstract}

\section{Introduction}\label{sec:introduction}
\input{sections/01_introduction}

\section{Related Work}\label{sec:related-work}
\input{sections/02_related_work}

\section{Source Data and Mask Construction}\label{sec:source-mask-construction}
\input{sections/03_01_source_and_mask_overview}
\input{sections/03_02_source_dataset}
\input{sections/03_03_cruise_extraction}

\section{Benchmark Definition}\label{sec:benchmark-definition}
\input{sections/04_01_task_formulation}
\input{sections/04_02_rul_target_construction}
\input{sections/04_03_windowing_and_temporal_downscaling}
\input{sections/04_04_evaluation_metrics}
\input{sections/04_05_main_benchmark_setting}

\section{Experimental Setup}\label{sec:experimental-setup}
\input{sections/05_01_baseline_models}
\input{sections/05_02_training_configuration_and_platform}

\section{Results and Ablations}\label{sec:results-ablations}
\input{sections/06_results_and_ablations}

\section{Discussion}\label{sec:discussion}
\input{sections/07_discussion}

\section{Conclusion}\label{sec:conclusion}
\input{sections/08_conclusion}

\section{Reproducibility}\label{sec:reproducibility}
\input{sections/09_reproducibility}

\section*{Acknowledgement}
This research was partially supported by the National Natural Science Foundation of China (No. 52075349) and Sichuan Science and Technology Program under Grant (No. 2025YFHZ0157).

\bibliographystyle{plainnat}
\bibliography{references}

\end{document}

%% file: sections/01_introduction.tex
Remaining useful life (RUL) prediction is a central task in prognostics and health management because maintenance planning depends on estimating how long monitored assets can continue safe operation. Aero-engines are a representative application. They are safety critical, heavily instrumented, and operated under changing environmental and control conditions. The original C-MAPSS benchmark provided simulated turbofan run-to-failure trajectories and became a standard testbed for data-driven prognostics \citep{saxenaDamagePropagationModeling2008}. N-CMAPSS extends this line of work by generating run-to-failure trajectories under real recorded flight profiles and operation-history-dependent degradation \citep{ariaschaoAircraftEngineRuntoFailure2021}.

The realism of N-CMAPSS also creates a benchmark-design issue. A complete flight cycle contains climb, cruise, and descent, and these stages differ clearly in altitude, Mach number, throttle-resolver angle, temperature, and sensor response. Full-flight records also increase data volume and make preprocessing choices more influential.
A model evaluated on full-cycle trajectories may therefore benefit from both degradation-sensitive temporal patterns extraction and operating-regime variation detection.
Cruise-stage benchmarking protocol can isolate one stable operating regime for controlled RUL model comparison.
It reduces variation caused by mixing flight stages while preserving a real-flight-aligned prediction task, thereby improving comparability across RUL models.

This paper proposes \benchmark, a cruise-stage benchmark derived from N-CMAPSS. It defines \datasetname\ (Cruising-Period Mask for N-CMAPSS), a reproducible mask artifact that stores accepted cruise intervals inside each flight cycle while retaining the original N-CMAPSS data splits and organization. The resulting evaluation setting focuses on cruise-stage RUL prediction, and the stage-specific mask can also support later transfer-learning and domain-adaptation studies across operating regimes. The ablation studies in Section~\ref{sec:ablation-studies} support this design by showing that flight-stage selection, temporal downscaling method, and RUL-cap threshold affect reported results. Together, these findings motivate treating flight stage, temporal downscaling, and cap threshold as explicit benchmark settings.

The contributions are:
\begin{itemize}[leftmargin=*]
    \item \datasetname, a mask artifact that stores reproducible cycle-local cruising intervals without modifying or redistributing the official raw files, with coverage across nine accessible subdatasets.
    \item \benchmark, a fixed cruise-stage RUL protocol that specifies the input feature groups, metadata exclusions, windowing rule, RUL-cap rule, and evaluation metrics.
    \item a stage-specific data foundation for future transfer-learning and domain-adaptation studies across operating regimes.
\end{itemize}

The rest of the paper is organized as follows. Section~\ref{sec:related-work} reviews related RUL benchmarks and protocol-comparability issues. Section~\ref{sec:source-mask-construction} describes the source data, mask artifact, and cruising-period extraction method. Section~\ref{sec:benchmark-definition} defines the supervised RUL protocol, including feature selection, RUL capping, windowing, and metrics. Section~\ref{sec:experimental-setup} describes the baseline models and training setup. Section~\ref{sec:results-ablations} reports the main results and ablations. Section~\ref{sec:discussion} discusses limitations. Section~\ref{sec:conclusion} summarizes the paper and future directions. Section~\ref{sec:reproducibility} documents reproducibility.

%% file: sections/02_related_work.tex
\subsection{C-MAPSS and N-CMAPSS benchmarks}
C-MAPSS established a widely used simulated turbofan degradation benchmark for RUL prediction \citep{saxenaDamagePropagationModeling2008}. It generates run-to-failure trajectories for a fleet of large turbofan engines by changing flow and efficiency parameters in the engine model. Each engine cycle is represented by one operating and sensor snapshot rather than by a complete within-flight time series. This representation greatly reduces data volume and makes algorithm comparison easier on C-MAPSS, but it does not fully describe flight-stage behavior inside a complete flight.

N-CMAPSS extends C-MAPSS by using real recorded flight profiles and operation-history-dependent degradation in a higher-fidelity simulation framework \citep{ariaschaoAircraftEngineRuntoFailure2021}. It simulates complete flights with climb, cruise, and descent stages, and stores development and test data in HDF5 files. Each file provides scenario descriptors $W$, measured sensors $X_s$, virtual sensors $X_v$, health parameters $\theta$, RUL labels, and auxiliary metadata such as unit, cycle, flight class, and health state. These additions make the benchmark more realistic, but they also introduce more choices in preprocessing and evaluation.

\subsection{Protocol Variation in N-CMAPSS Studies}
Reported N-CMAPSS results differ not only by model architecture, but also by protocol choices. Table~\ref{tab:ncmapss-protocol-comparison} summarizes representative and latest choices reported in N-CMAPSS RUL studies. The table concludes subdataset coverage, temporal reduction, window stride, target capping, and feature groups in columns.

\begin{table}[h!]
\centering
\scriptsize
\setlength{\tabcolsep}{2.5pt}
\renewcommand{\arraystretch}{1.08}
\caption{Representative N-CMAPSS RUL protocol choices in recent work. The sample-rate factor is relative to the native N-CMAPSS time resolution. NR means not reported.}
\label{tab:ncmapss-protocol-comparison}
\begin{tabular}{>{\raggedright\arraybackslash}p{0.16\linewidth}
                >{\raggedright\arraybackslash}p{0.15\linewidth}
                >{\raggedright\arraybackslash}p{0.08\linewidth}
                >{\raggedright\arraybackslash}p{0.12\linewidth}
                >{\raggedright\arraybackslash}p{0.11\linewidth}
                >{\raggedright\arraybackslash}p{0.25\linewidth}}
\toprule
Study & Datasets used & Sample-rate factor & Window size / stride & RUL cap & Input channels or feature rule \\
\midrule
\citet{ariaschaoFusingPhysicsbasedDeep2022} & DS02 & x10 & 50 / 1 & NR & $W+X_s$, hybrid adds $\hat{X}_s+\hat{X}_v+\theta$ \\
\citet{custodeEvolutionaryOptimizationSpiking2022} & DS02 & x100 & NR & NR & 20 condition-monitoring signals \\
\citet{zengDeepGaussianProcess2023a} & DS01--DS07, DS08a, DS08c & x100 & 30 / 1 & No early cap & $W+X_s+X_v$ \\
\citet{xuMultiscaleBLSBasedLightweight2024a} & DS01--DS07 & NR & NR & NR & Zero-valued N-CMAPSS variables removed, retained group NR \\
\citet{wang2024gpt} & DS01--DS07 & NR & 40 / 1 & 65 & $W+X_s$ \\
\citet{songEnhancingRemainingUseful2025} & DS02 & x100 & 100, 200 / NR & NR & $W$ for clustering, sensor channels for graph learning \\
\citet{forestInterpretablePrognosticsConcept2025} & DS01, DS04, DS05, DS07 & x10 & NR & Flat or linear RUL option & Concept-bottleneck inputs with degradation-mode concepts \\
\citet{wuDualpathArchitectureBased2025} & DS02 & x10 & 60 / 1 & NR & $X_s+X_v+\theta$, seven constant $\theta$ columns removed \\
\citet{zhangNovelLocalEnhanced2025} & DS02 & x1000 & 70 / 1 & NR & $W+X_s+X_v$ \\
\citet{landauFederatedLearningFramework2026} & DS02 & x20 & 50 / 10 & NR & $W+$ selected $X_s$ \\
\citet{qinMultisensorSpatialMultiscale2026} & DS02 & x1 & 60 / 1 & 120 & $W+X_s$ \\
\citet{renSpatialtemporalGraphHybrid2026} & DS01--DS07, DS08a & x100 & 2 / 1 & NR & N-CMAPSS feature group NR \\
\citet{zhuEfficientTemporalAdjacent2026} & DS02 & x10 & NR, patch 16 / 8 & NR & $X_s$ \\
\citet{gaoVSCNetVersatileSpatiotemporal2025} & Subset NR & NR & NR & NR & N-CMAPSS feature group NR \\
\citet{zhouAeroengineRemainingUseful2026} & Subset NR & NR & NR & NR & N-CMAPSS feature group NR \\
\benchmark\ (this work) & DS01--DS07, DS08a, DS08c & x1 & 256 / 10 & Dataset-wise $\eta=5\%$ caps & $W+X_s$ \\
\bottomrule
\end{tabular}
\end{table}

The input-feature construction is one important source of variation. The original N-CMAPSS data-driven setting uses scenario descriptors $W$ and measured sensors $X_s$ as inputs \citep{ariaschaoAircraftEngineRuntoFailure2021}. Arias Chao et al. compare this data-driven rule with hybrid variants that add calibrated sensor estimates $\hat{X}_s$, virtual sensors $\hat{X}_v$, and health parameters $\theta$ \citep{ariaschaoFusingPhysicsbasedDeep2022}. Other studies in Table~\ref{tab:ncmapss-protocol-comparison} use combinations such as $W+X_s+X_v$, $X_s+X_v+\theta$, selected sensor subsets, or concept variables. These settings can be appropriate for their respective objectives, but inconsistent sensor selection complicates benchmark construction and direct comparison.

The N-CMAPSS subdataset choice is also part of the experimental setting. DS02 is one of the N-CMAPSS datasets used in prior prognostics studies \citep{ariaschaoAircraftEngineRuntoFailure2021}. It contains nine run-to-failure units across flight classes 1, 2, and 3, includes complete flight traces and several variable groups, and is useful for studies beyond a simple data-driven RUL setting. However, DS02-only evaluation can narrow the benchmark scope. As Table~\ref{tab:ncmapss-protocol-comparison} shows, several studies report DS02 only \citep{ariaschaoFusingPhysicsbasedDeep2022,landauFederatedLearningFramework2026,qinMultisensorSpatialMultiscale2026,wuDualpathArchitectureBased2025,zhangNovelLocalEnhanced2025}. Other studies use selected groups, such as DS02, DS05, DS06, and DS07, or DS01--DS07 \citep{liuMultiScaleTemporalSpatial2024,xuMultiscaleBLSBasedLightweight2024a,wang2024gpt}. A nine-subdataset setting, namely DS01--DS07, DS08a, and DS08c, is also used after excluding DS08d \citep{zengDeepGaussianProcess2023a,huangReScConvXLSTM2025}. Ren et al. instead evaluate DS01--DS07 and DS08a \citep{renSpatialtemporalGraphHybrid2026}. These examples show that subdataset coverage is part of the reported setting rather than a fixed community default.

Preprocessing choices introduce another source of variation. Because N-CMAPSS contains millions of samples in each subset, studies may reduce the training and testing burden through temporal downscaling methods such as sample-rate reduction or window-stride settings. For example, Liu et al. downsample the selected data for this reason \citep{liuMultiScaleTemporalSpatial2024}. Among the studies summarized in Table~\ref{tab:ncmapss-protocol-comparison}, reported sample-rate factors range from native resolution to x1000, while window lengths, strides, and patch sizes also vary or are left unreported.
The table also shows that target construction is not uniform. RUL caps may be not used, fixed to a stated value, dataset-wise, or not reported. Reported results can further depend on whether health-state or flight-class metadata are used, the metric definition, and whether a number is selected from one run or averaged across repeated runs. These differences make direct comparison difficult, and results reported under different procedures should not be treated as one unified leaderboard.

The last row of Table~\ref{tab:ncmapss-protocol-comparison} states that the proposed \benchmark\ fixes a cruise-stage setting across the nine accessible subdatasets, native sample rate, 256-sample windows with stride 10, dataset-wise label caps, and model inputs restricted to the original data-driven groups $W+X_s$. Making these choices explicit helps distinguish gains from model design from gains caused by preprocessing, feature selection, or target construction.

%% file: sections/03_01_source_and_mask_overview.tex
\datasetname\ (Cruising-Period Mask for N-CMAPSS) is a cruise-stage mask artifact for N-CMAPSS. N-CMAPSS contains complete run-to-failure aero-engine trajectories generated by using recorded flight profiles as inputs to the C-MAPSS turbofan model and by imposing degradation through component health parameters \citep{ariaschaoAircraftEngineRuntoFailure2021}. In \datasetname, each accepted cruising period is stored as interval metadata in the local sample coordinates of one unit-cycle trace. Each entry records the dataset split, unit, cycle, and start--end sample indices needed to select the corresponding official HDF5 rows. Benchmark construction applies this mask to obtain the cruise-stage subset while leaving the source files unchanged. This section describes the source-file structure, the mask artifact, and the cruising-period extraction method.

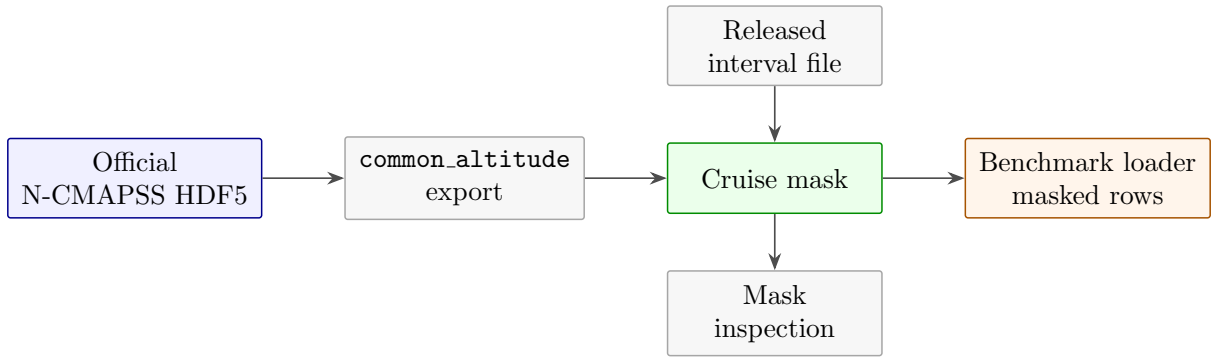
\begin{figure}[H]
\centering
\resizebox{\linewidth}{!}{%
\begin{tikzpicture}[
    font=\small,
    sourcebox/.style={draw=blue!55!black, fill=blue!6, rounded corners=1pt, line width=0.55pt, align=center, inner sep=5pt, minimum width=2.75cm, minimum height=0.9cm},
    maskbox/.style={draw=green!55!black, fill=green!8, rounded corners=1pt, line width=0.55pt, align=center, inner sep=5pt, minimum width=2.75cm, minimum height=0.9cm},
    toolbox/.style={draw=gray!70, fill=gray!6, rounded corners=1pt, line width=0.55pt, align=center, inner sep=5pt, minimum width=2.75cm, minimum height=0.9cm},
    loaderbox/.style={draw=orange!65!black, fill=orange!8, rounded corners=1pt, line width=0.55pt, align=center, inner sep=5pt, minimum width=2.75cm, minimum height=0.9cm},
    arrow/.style={-{Stealth[length=2.2mm]}, line width=0.6pt, draw=black!70},
    node distance=1.05cm
]
\node[sourcebox] (hdf5) {Official\\N-CMAPSS HDF5};
\node[toolbox, right=of hdf5] (export) {\texttt{common\_altitude}\\export};
\node[maskbox, right=of export] (mask) {Cruise mask};
\node[loaderbox, right=of mask] (loader) {Benchmark loader\\masked rows};
\node[toolbox, above=0.72cm of mask] (reuse) {Released\\interval file};
\node[toolbox, below=0.72cm of mask] (check) {Mask\\inspection};

\draw[arrow] (hdf5) -- (export);
\draw[arrow] (export) -- (mask);
\draw[arrow] (reuse) -- (mask);
\draw[arrow] (mask) -- (loader);
\draw[arrow] (mask) -- (check);
\end{tikzpicture}%
}
\caption{\datasetname\ mask workflow. The mask can be reproduced from official N-CMAPSS HDF5 files or reused from the released interval file. Inspection tools check the same artifact before the benchmark loader applies it to source rows.}
\label{fig:mask-workflow}
\end{figure}

\subsection{Mask Artifact}
The HDF5 mask file stores interval metadata for accepted unit-cycle pairs. Each interval is expressed by local sample indices inside one flight cycle. This design keeps the artifact small, avoids duplicating N-CMAPSS sensor arrays, and makes the mask auditable with the original altitude trace.

\subsection{Open-source Mask Workflow}
The open-source package supports both mask reproduction and direct reuse of the released mask, as summarized in Figure~\ref{fig:mask-workflow}. Users place official N-CMAPSS files in the local data directory, run the common-altitude export tool when reproduction is needed, and inspect the generated intervals. During training, the data loader applies the checked mask to the official source rows and then follows the benchmark rules defined in Section~\ref{sec:benchmark-definition}.

%% file: sections/03_02_source_dataset.tex
\subsection{Source Files and Cruise Mask}
The source data are the N-CMAPSS HDF5 files. Each dataset split provides scenario descriptors $W$, measured sensors $X_s$, virtual sensors $X_v$, auxiliary metadata $A$, health parameters $\theta$, and RUL labels. The auxiliary metadata include unit id, cycle id, flight class, sample index, and health state. \datasetname\ covers nine accessible subdatasets of N-CMAPSS: DS01-005, DS02-006, DS03-012, DS04, DS05, DS06, DS07, DS08a-009, and DS08c-008. DS08d-010 is excluded because its available HDF5 file could not be read. Table~\ref{tab:mask-summary} summarizes the released mask file.

\begin{table}[H]
\centering
\caption{\datasetname\ cruise-mask export summary. Cycle ratio is accepted cycles divided by total cycles. Sample ratio is accepted cruise samples divided by total samples.}
\label{tab:mask-summary}
\begin{tabular}{lrrrr}
\toprule
Dataset & Cycles & Accepted cycles & Cycle ratio & Sample ratio \\
\midrule
DS01-005 & 894 & 835 & 0.934 & 0.264 \\
DS02-006 & 648 & 611 & 0.943 & 0.340 \\
DS03-012 & 1101 & 1026 & 0.932 & 0.259 \\
DS04 & 856 & 827 & 0.966 & 0.252 \\
DS05 & 818 & 768 & 0.939 & 0.273 \\
DS06 & 797 & 747 & 0.937 & 0.273 \\
DS07 & 812 & 762 & 0.938 & 0.268 \\
DS08a-009 & 994 & 943 & 0.949 & 0.261 \\
DS08c-008 & 553 & 534 & 0.966 & 0.267 \\
\midrule
Total & 7473 & 7053 & 0.944 & 0.271 \\
\bottomrule
\end{tabular}
\end{table}

%% file: sections/03_03_cruise_extraction.tex
\subsection{Cruising-Period Extraction}\label{sec:cruise-extraction}
Cruising-period extraction assigns cycle-local sample indices to the part of a flight cycle selected as cruise. Let a cycle contain $n$ samples indexed by $i=0,1,\ldots,n-1$, and let $a_i$ denote the raw altitude at sample $i$. The released mask uses a fixed common-altitude method, denoted by
\begin{equation}
    \mathcal{C}_{\theta}: (a_0,\ldots,a_{n-1})
    \mapsto \mathcal{I}_c,
\end{equation}
where $\theta$ denotes the fixed method parameters. The output $\mathcal{I}_c$ is either a singleton set $\{(s,e)\}$ with $0 \leq s \leq e < n$ or an empty set when no cruising period is accepted.
The exported interval is expanded into a binary sample when loaded:

\begin{equation}
    m_i =
    \begin{cases}
        1, & \text{if } \mathcal{I}_c=\{(s,e)\} \text{ and } s \leq i \leq e,\\
        0, & \text{otherwise.}
    \end{cases}
\end{equation}

\subsubsection{Common-altitude method}\label{sec:common-altitude-method}
The common-altitude method searches each cycle for a sustained high-altitude level segment. It first smooths altitude with a 31-sample centered rolling median:
\begin{equation}
    \tilde{a}_i = \mathrm{median}\{a_j : j \in \mathcal{W}_i\},
\end{equation}
where
\begin{equation}
    \mathcal{W}_i =
    \{j: 0 \leq j < n,\ |j-i| \leq 15\}.
\end{equation}
The cycle is processed only when its smoothed altitude span is large enough:
\begin{equation}
    S = \max_i \tilde{a}_i - \min_i \tilde{a}_i .
\end{equation}
If $S < 500$ ft, no cruise interval is exported for that cycle.

Next, the method defines a lower altitude floor:
\begin{equation}
    F = \max\left(Q_{0.50}(\tilde{a}), \min_i \tilde{a}_i + 0.35S\right).
\end{equation}
Here, $Q_{0.50}(\tilde{a})$ is the median of the smoothed altitude sequence. Only candidate altitude levels at or above $F$ are considered.

The method then searches for repeated high-altitude levels using shifted bins. Let $\Delta=50$ ft be the bin width and $K=50$ be the number of bin shifts. For shift $k=0,\ldots,K-1$, the bin origin is $o_k=k\Delta/K$, and sample $i$ is assigned to the bin center
\begin{equation}
    b_i^{(k)} =
    \left\lfloor \frac{\tilde{a}_i-o_k}{\Delta} \right\rfloor \Delta
    + o_k + \frac{\Delta}{2}.
\end{equation}
For each shift, the method selects the three most frequent bin centers among samples with $b_i^{(k)} \geq F$. For a selected candidate level $\lambda$, the initial candidate mask is
\begin{equation}
    r_i(\lambda,k) =
    \begin{cases}
        1, & \text{if } |\tilde{a}_i-\lambda| \leq 50,\\
        0, & \text{otherwise.}
    \end{cases}
\end{equation}
Short false gaps of at most 80 samples are merged when the adjacent smoothed altitude values still belong to the same plateau. The method then extracts continuous true runs from the repaired candidate mask. A run from $s$ to $e$ is accepted only if
\begin{equation}
    e-s+1 \geq 512
    \quad \text{and} \quad
    \mathrm{std}\{a_i: s \leq i \leq e\} \leq 50\ \mathrm{ft}.
\end{equation}
Among all accepted runs from all shifts and candidate levels, the mask export keeps the longest one and writes its endpoints in cycle-local sample indices. If no run satisfies these criteria, no cruise interval is exported for that cycle.

\subsubsection{Extraction examples}\label{sec:cruise-extraction-examples}
\begin{figure}[H]
\centering
\includegraphics[width=\linewidth]{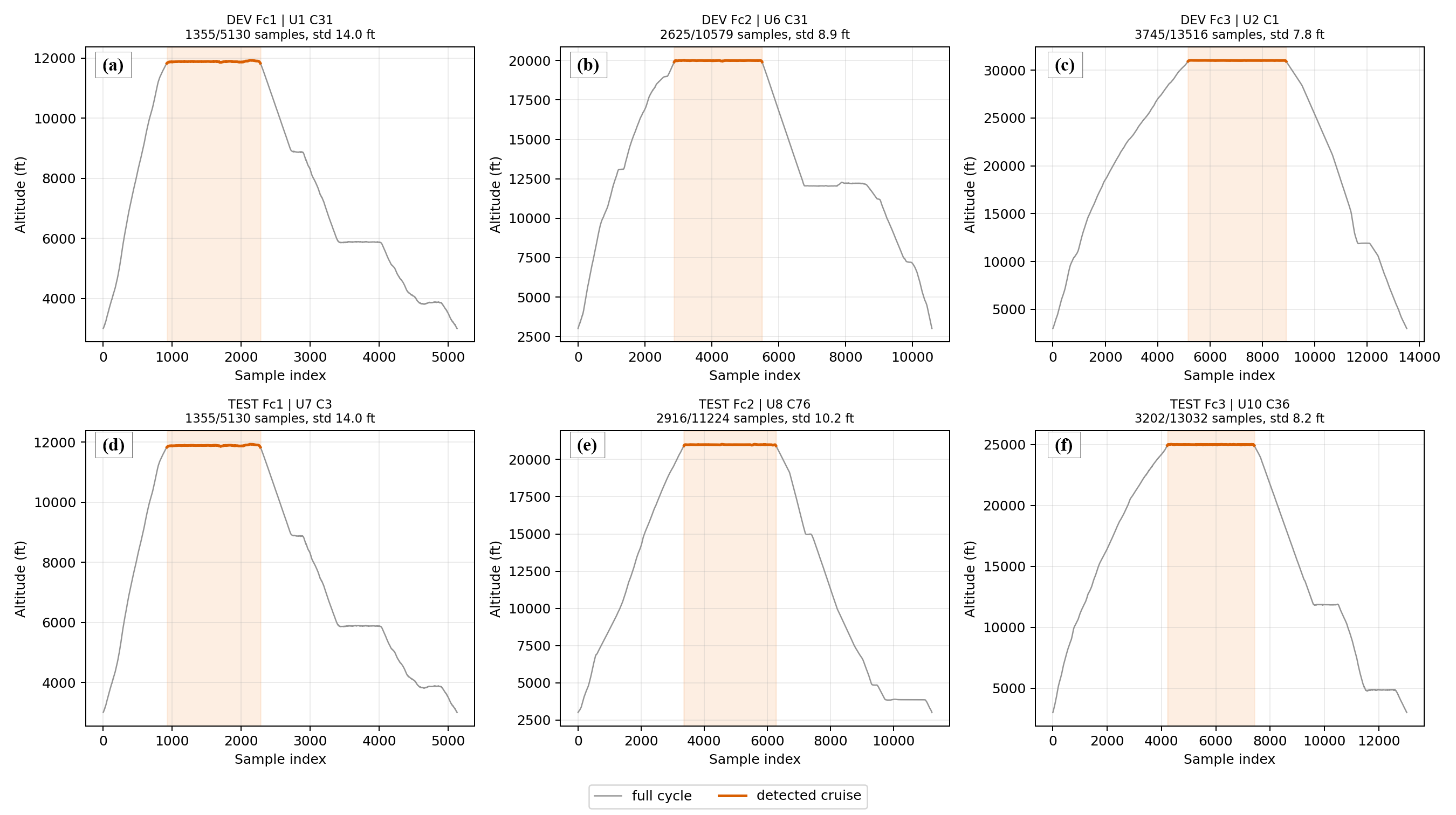}
\caption{Accepted cruising-period examples from DS01-005. The mask is generated cycle by cycle from altitude and then exported as cycle-local start and end sample indices.}
\label{fig:cruise-example}
\end{figure}

Figure~\ref{fig:cruise-example} shows representative accepted cruising periods from DS01-005. The selected interval is generated from altitude within each cycle and exported as cycle-local start and end sample indices.

\begin{figure}[H]
\centering
\includegraphics[width=\linewidth]{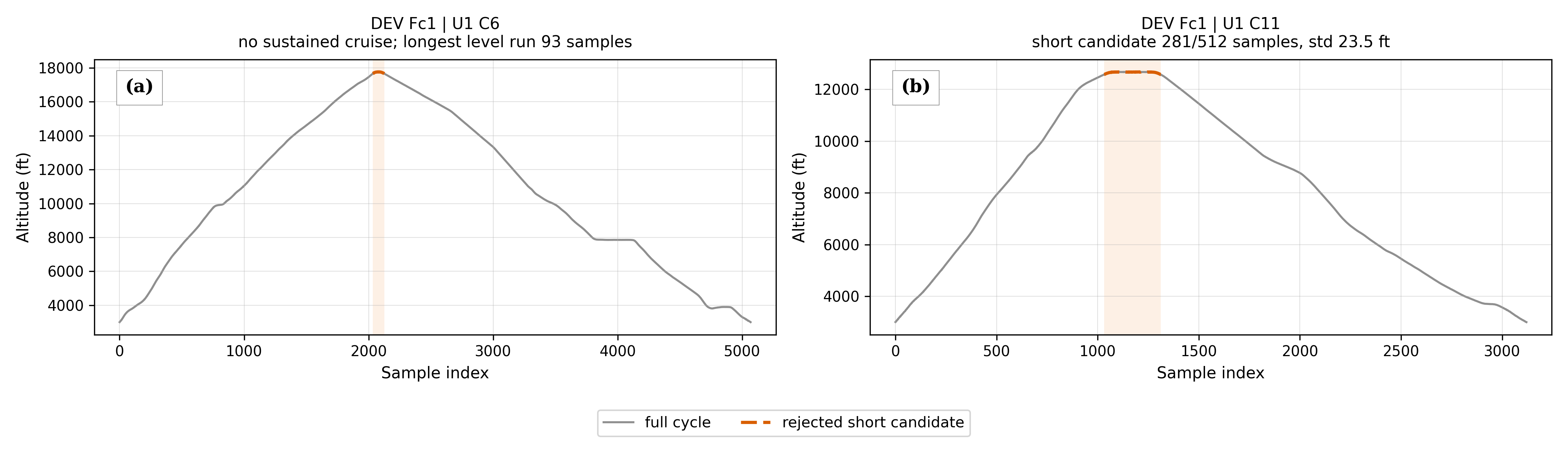}
\caption{Examples of DS01-005 cycles skipped by the common-altitude method. Orange dashed segments show the longest candidate level run found before the minimum-duration test. (a) Unit 1 Cycle 6 does not form a sustained cruise-level plateau; its longest level run is 93 samples. (b) Unit 1 Cycle 11 contains a short stable candidate, but 281 samples is below the 512-sample selection threshold.}
\label{fig:cruise-skipped-example}
\end{figure}

Cycles that do not contain a sustained common-altitude segment are rejected. Figure~\ref{fig:cruise-skipped-example} shows two DS01-005 development examples: Unit 1 Cycle 6 has no sustained high-altitude plateau, while Unit 1 Cycle 11 contains a visually plausible level segment that lasts only 281 samples, below the 512-sample minimum. These rejected cycles receive no fallback interval and are omitted at the cycle-local matching step.

%% file: sections/04_01_task_formulation.tex
\subsection{Task Formulation}

\benchmark\ follows the N-CMAPSS prognostics problem definition introduced by Arias Chao et al.~\citep{ariaschaoAircraftEngineRuntoFailure2021}. Given a time window of operating and sensor measurements, the model predicts the remaining useful life of the engine. \benchmark\ keeps this supervised RUL regression formulation and applies it to the cruise-stage rows selected by the mask.
The model input includes only the scenario descriptors $W$ and measured sensors $X_s$. In the official N-CMAPSS files, $W$ provides four channels and $X_s$ provides fourteen channels. Let $T$ denote the window length and let $d=18$ denote the number of input channels. A windowed input therefore has shape
\begin{equation}
    X_j \in \mathbb{R}^{T \times d}.
\end{equation}
Virtual sensors $X_v$, health parameters $\theta$, and auxiliary metadata $A$ are not used as predictive features. Metadata fields such as unit id, cycle id, flight class, sample index, and health state are used only for filtering or offline setting design.

For a model $f_\phi$, the benchmark prediction for window $X_j$ is
\begin{equation}
    \hat{y}_j = f_\phi(X_j).
\end{equation}

%% file: sections/04_02_rul_target_construction.tex
\subsection{RUL Target Construction}

Cap is applied to the original N-CMAPSS RUL labels to form the supervised target. RUL capping is a common target-labeling step in C-MAPSS studies: the target is kept constant during the early-life part and then follows the true RUL near failure \citep{heimesRecurrentNeuralNetworks2008,liRemainingUsefulLife2018,ellefsenValidationDatadrivenLabeling2019}. This makes the training target emphasize the late-life region where degradation is clearer. Many C-MAPSS comparisons use a single cap, often 125 cycles, because the benchmark is organized around cycle-level trajectories with a relatively compact labeling rule.

N-CMAPSS requires a dataset-wise rule. It contains complete flight traces, multiple subdatasets, and health-state metadata generated by the simulator. Health state is not used as a model input in \benchmark, but it can be used offline to define dataset-wise target caps. Therefore caps are derived from the exponential abnormal-degradation law and treat the threshold $\eta$ as part of the benchmark setting rather than as a directly observed detectability limit. Following the N-CMAPSS abnormal degradation model \citep{ariaschaoAircraftEngineRuntoFailure2021}, offset and noise are ignored and the abnormal damage magnitude is written as
\begin{equation}
    D(t) = \exp(a t^b)-1,
\end{equation}
with $a \in [0.001,0.003]$ and $b \in [1.4,1.6]$. For dataset $i$ with minimum abnormal length $L_i$, define a normalized curve
\begin{equation}
    \bar{D}_i(u) =
    \frac{\exp(a L_i^b u^b)-1}{\exp(a L_i^b)-1},
    \quad u \in [0,1].
\end{equation}
Given a detectable-damage threshold $\eta$, solve $\bar{D}_i(u_\eta)=\eta$ and set
\begin{equation}
    \mathrm{MAX\_RUL}_i =
    \mathrm{round}(L_i(1-u_\eta)).
\end{equation}
The regression target for a window from dataset $i$ is therefore
\begin{equation}
    y = \min(\mathrm{RUL}, \mathrm{MAX\_RUL}_i).
\end{equation}
This transformation keeps the late-life target unchanged while mapping earlier windows to a dataset-specific plateau. Table~\ref{tab:rul-cap-settings} lists the minimum abnormal length used for each dataset and the resulting caps for the main setting and the two sensitivity settings. The main setting uses $a=0.002$, $b=1.5$, and $\eta=5\%$. The $\eta=10\%$ setting uses a stricter cap, while the $\eta=0\%$ setting uses $L_i$ itself as the dataset-wise cap. The three settings are reported separately when used.

\begin{table}[H]
\centering
\caption{Minimum abnormal lengths and dataset-wise RUL caps. The main setting uses the $\eta=5\%$ cap.}
\label{tab:rul-cap-settings}
\small
\begin{tabular}{lrrrr}
\toprule
Dataset & Minimum abnormal length $L_i$ & $\eta=0\%$ cap & $\eta=5\%$ cap & $\eta=10\%$ cap \\
\midrule
DS01-005 & 46 & 46 & 38 & 34 \\
DS02-006 & 41 & 41 & 34 & 31 \\
DS03-012 & 41 & 41 & 34 & 31 \\
DS04 & 52 & 52 & 43 & 38 \\
DS05 & 46 & 46 & 38 & 34 \\
DS06 & 50 & 50 & 41 & 37 \\
DS07 & 47 & 47 & 39 & 35 \\
DS08a-009 & 40 & 40 & 34 & 30 \\
DS08c-008 & 33 & 33 & 28 & 25 \\
\bottomrule
\end{tabular}
\end{table}

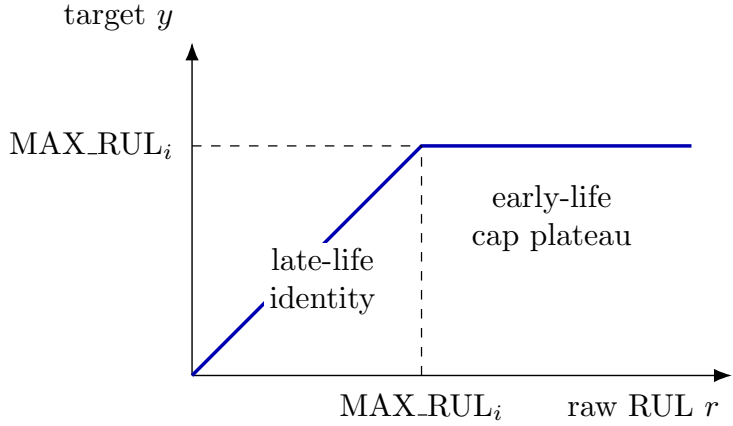
\begin{figure}[H]
\centering
\resizebox{0.62\linewidth}{!}{%
\begin{tikzpicture}[
    font=\small,
    axis/.style={-{Latex[length=2mm]}, line width=0.45pt},
    guide/.style={dashed, line width=0.4pt},
    plotlabel/.style={align=center, fill=white, inner sep=1.5pt}
]
\draw[axis] (0,0) -- (6.0,0) node[below=2pt, anchor=north east] {raw RUL $r$};
\draw[axis] (0,0) -- (0,3.7) node[left=2pt, anchor=south east] {target $y$};
\draw[blue!70!black, line width=1.1pt] (0,0) -- (2.55,2.55) -- (5.55,2.55);
\draw[guide] (2.55,0) -- (2.55,2.55);
\draw[guide] (0,2.55) -- (2.55,2.55);
\node[below=2pt] at (2.55,0) {$\mathrm{MAX\_RUL}_i$};
\node[left=2pt] at (0,2.55) {$\mathrm{MAX\_RUL}_i$};
\node[plotlabel, anchor=east] at (2.10,1.05) {late-life\\identity};
\node[plotlabel, anchor=west] at (3.05,1.75) {early-life\\cap plateau};
\end{tikzpicture}
}
\caption{RUL target transformation after dataset-wise capping. Late-life RUL values remain unchanged, while earlier windows are mapped to a dataset-specific plateau.}
\label{fig:rul-capping}
\end{figure}

%% file: sections/04_03_windowing_and_temporal_downscaling.tex
\subsection{Windowing and Temporal Downscaling}

The main setting uses $T=256$ native-resolution samples per window. This window size is chosen relative to the cruise-mask method. Because the method requires at least 512 cruise samples, a 256-sample window can fit within a minimum accepted cruise segment while still covering a substantial cruise-stage context.

The benchmark uses window stride as the temporal downscaling method and evaluates sample-rate downsampling in the ablation. Both methods reduce temporal density. The main setting fixes native-resolution windows with stride, while the ablation compares this choice with sample-rate downsampling. The main setting uses stride $s=10$ at the original N-CMAPSS sample rate. This setting changes how often window starts are selected, but it keeps the native time interval between neighboring samples inside each window. With window size $T$ and stride $s$, the $j$-th native-resolution window is
\begin{equation}
    X_j = [x_{1+js}, x_{2+js}, \ldots, x_{T+js}],
\end{equation}
where $j=0,1,2,\ldots$. Increasing $s$ therefore reduces overlapping, near-duplicate windows while preserving the temporal content of each input sequence.

The difference can be seen from the number of usable windows. For an accepted cruise interval with $N$ native samples, stride-only windowing gives
\begin{equation}
    M_{\mathrm{stride}}(N;s) =
    \begin{cases}
        \left\lfloor \dfrac{N-T}{s} \right\rfloor + 1, & N \ge T,\\
        0, & N < T.
    \end{cases}
\end{equation}

If the signal is first downsampled in sample rate by a factor $k$ while the $T$-sample window length is kept unchanged, the reduced sequence contains approximately $\lfloor N/k \rfloor$ samples. The corresponding number of windows is
\begin{equation}
    M_{\mathrm{sample}}(N;k) =
    \begin{cases}
        \left\lfloor \dfrac{N}{k} \right\rfloor - T + 1, & \left\lfloor \dfrac{N}{k} \right\rfloor \ge T,\\
        0, & \left\lfloor \dfrac{N}{k} \right\rfloor < T.
    \end{cases}
\end{equation}

\begin{figure}[H]
\centering
\includegraphics[width=0.8\linewidth]{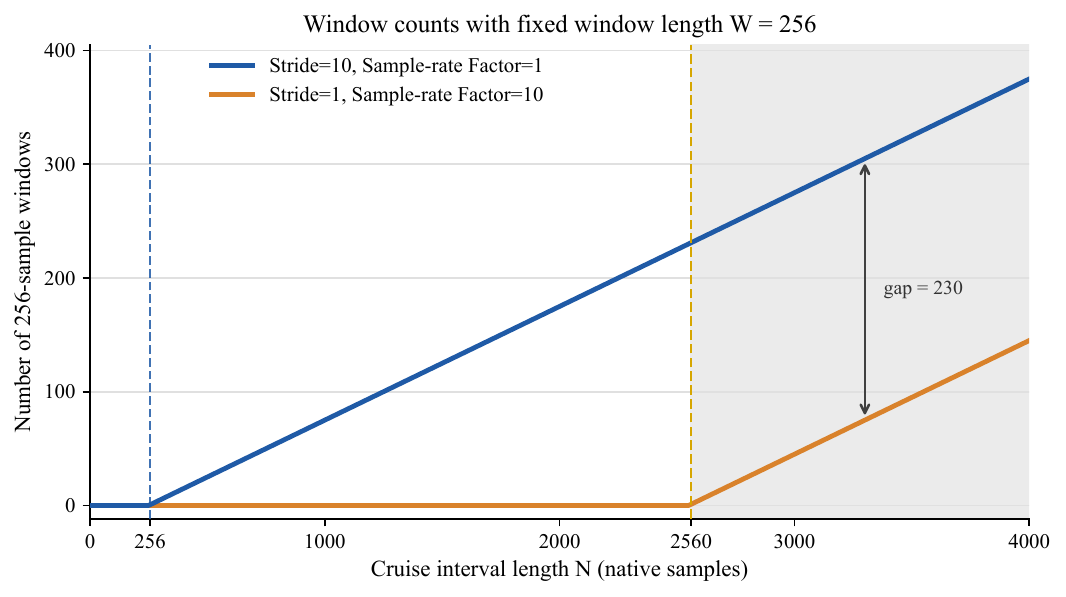}
\caption{Window-count comparison for fixed window length $T=256$. The blue dashed line marks the first native-resolution window at $N=256$, and the yellow dashed line marks the first downsampled window at $N=2560$. The shaded region marks where the window-count gap between the two settings is constant at 230. Counts are evaluated every ten native samples.}
\label{fig:stride-sample-window-counts}
\end{figure}

Thus, window stride and sample-rate downsampling produce inputs with the same tensor shape, but the samples inside each input represent different time scales and different temporal detail.
With $T=256$ and $s=k=10$, a 2560-sample cruise interval yields 231 native-resolution stride windows but only one downsampled window. On the ten-sample length grid shown in Figure~\ref{fig:stride-sample-window-counts}, both counts then increase by one for every additional ten native samples, so the gap remains 230. Accordingly, \benchmark\ uses stride as the benchmark parameter for downscaling and keeps the sample-rate factor fixed at one in the main setting.

%% file: sections/04_04_evaluation_metrics.tex
\subsection{Evaluation Metrics}

The supervised loss is the mean squared error between the prediction $\hat{y}_j$ and the capped RUL target $y_j$. Reported accuracy uses RMSE in cycles:
\begin{equation}
    \mathrm{RMSE} =
    \sqrt{\frac{1}{N}\sum_{j=1}^{N}(\hat{y}_j-y_j)^2}.
\end{equation}
For compatibility with C-MAPSS-style RUL reporting, the asymmetric score proposed by Saxena et al. is adopted~\citep{saxenaDamagePropagationModeling2008}. Let
\begin{equation}
    \Delta_j = y_j-\hat{y}_j .
\end{equation}
The score is
\begin{equation}
    S_{\mathrm{Saxena}} =
    \sum_{j=1}^{N}\left(\exp(\alpha_j|\Delta_j|)-1\right),
    \quad
    \alpha_j =
    \begin{cases}
        1/13, & \text{if } \Delta_j>0 \text{, i.e., RUL is under-estimated},\\
        1/10, & \text{otherwise.}
    \end{cases}
\end{equation}
Lower values are better for both RMSE and the Saxena score. The latter penalizes over-estimation more strongly because delayed maintenance predictions are more risky.

%% file: sections/04_05_main_benchmark_setting.tex
\subsection{Main Benchmark Setting}

The main benchmark setting is \mainsetting. It applies the released \datasetname\ mask generated by the common-altitude method and follows the task, target, windowing, and metric definitions above. Its fixed configuration is summarized in Table~\ref{tab:main-benchmark-setting}.

\begin{table}[H]
\centering
\caption{Summary of the main benchmark setting.}
\label{tab:main-benchmark-setting}
\small
\begin{tabular}{ll}
\toprule
Setting & Value \\
\midrule
Cruise-mask artifact & \datasetname \\
Cruising-period method & common-altitude \\
Input feature groups & $W$, $X_s$ \\
Excluded feature groups & $X_v$, $\theta$, $A$ \\
RUL-cap threshold $\eta$ & 5\% \\
Window size $T$ & 256 \\
Stride $s$ & 10 \\
Sample-rate factor & 1 \\
Evaluation metrics & RMSE, Saxena score \\
\bottomrule
\end{tabular}
\end{table}

%% file: sections/05_01_baseline_models.tex
\subsection{Baseline Models}
The baseline evaluation uses four standard sequence models: LSTM, GRU, TCN, and TSMixer. These models represent recurrent, convolutional, and MLP-based temporal encoders. All baselines follow the benchmark input and output rules defined in Section~\ref{sec:benchmark-definition}. The experiments provide reference performance under \benchmark\ protocol. Table~\ref{tab:model-hyperparameters} lists the model hyperparameters, which are shared across all datasets in the main setting.

\begin{table}[H]
\centering
\caption{Baseline model hyperparameters. All models use dropout 0.1, default parameter initialization, and a linear regression head that outputs one scalar RUL value.}
\label{tab:model-hyperparameters}
\small
\begin{tabular}{lllll}
\toprule
Model & Layers/blocks & Hidden or FFN dim. & Kernel/dilation \\
\midrule
LSTM & 2 & 32 & -- \\
GRU & 2 & 32 & -- \\
TCN & 2 & 32 & $k=3$, dilations 1 and 2 \\
TSMixer & 2 & 36 & time mixer and series mixer \\
\bottomrule
\end{tabular}
\end{table}

%% file: sections/05_02_training_configuration_and_platform.tex
\subsection{Training Setting and Platform}
The baseline experiments use a fixed training configuration across models and datasets. Training minimizes mean squared error. Evaluation uses RMSE and the Saxena asymmetric score. Table~\ref{tab:training-setting} lists the training settings used for the main run grid. The experiments were run on a workstation with an AMD Ryzen Threadripper PRO 5965WX CPU, 130 GB memory, and three NVIDIA GeForce RTX 4090 GPUs.

\begin{table}[H]
\centering
\caption{Training settings used in the baseline experiments.}
\label{tab:training-setting}
\small
\begin{tabular}{ll}
\toprule
Item & Value \\
\midrule
Epochs & 50 \\
Train batch size & 512 \\
Test batch size & 512 \\
Optimizer & Adam \\
Effective weight decay & 0 \\
Learning rate & $3\times10^{-4}$ \\
Scheduler & None \\
Loss & MSE \\
Repeated runs & Five random seeds \\
Distributed training & 3 GPUs \\
Mixed precision & No \\
\bottomrule
\end{tabular}
\end{table}

%% file: sections/06_results_and_ablations.tex
\subsection{Main Results}
The experiment covers the nine accessible subdatasets DS01-005, DS02-006, DS03-012, DS04, DS05, DS06, DS07, DS08a-009, and DS08c-008. DS08d remains excluded because its available HDF5 file could not be read. For each run, the best test RMSE is recorded, and be averaged across the five repeated runs. The main comparison uses \mainsetting. The $\eta=0\%$ and $\eta=10\%$ variants are reported as well.

\input{tables/benchmark_results_table}

Table~\ref{tab:benchmark-results} shows that the cruise-only setting retains learnable RUL structure for all baseline families, while difficulty remains dataset-dependent. Averaging the dataset-specific entries under the main $\eta=5\%$ setting, TSMixer obtains the lowest mean RMSE, $3.46\pm1.71$, and the lowest Saxena score, $(2.50\pm2.99)\times10^4$. TCN follows with $3.89\pm1.81$ RMSE and $(2.92\pm2.92)\times10^4$ score. This supports the benchmark argument. Ordinary sequence models can produce meaningful predictions in the fixed cruise-stage regime, but model choice still changes accuracy and stability under the same data setting.

\begin{figure}[H]
\centering
\includegraphics[width=\linewidth]{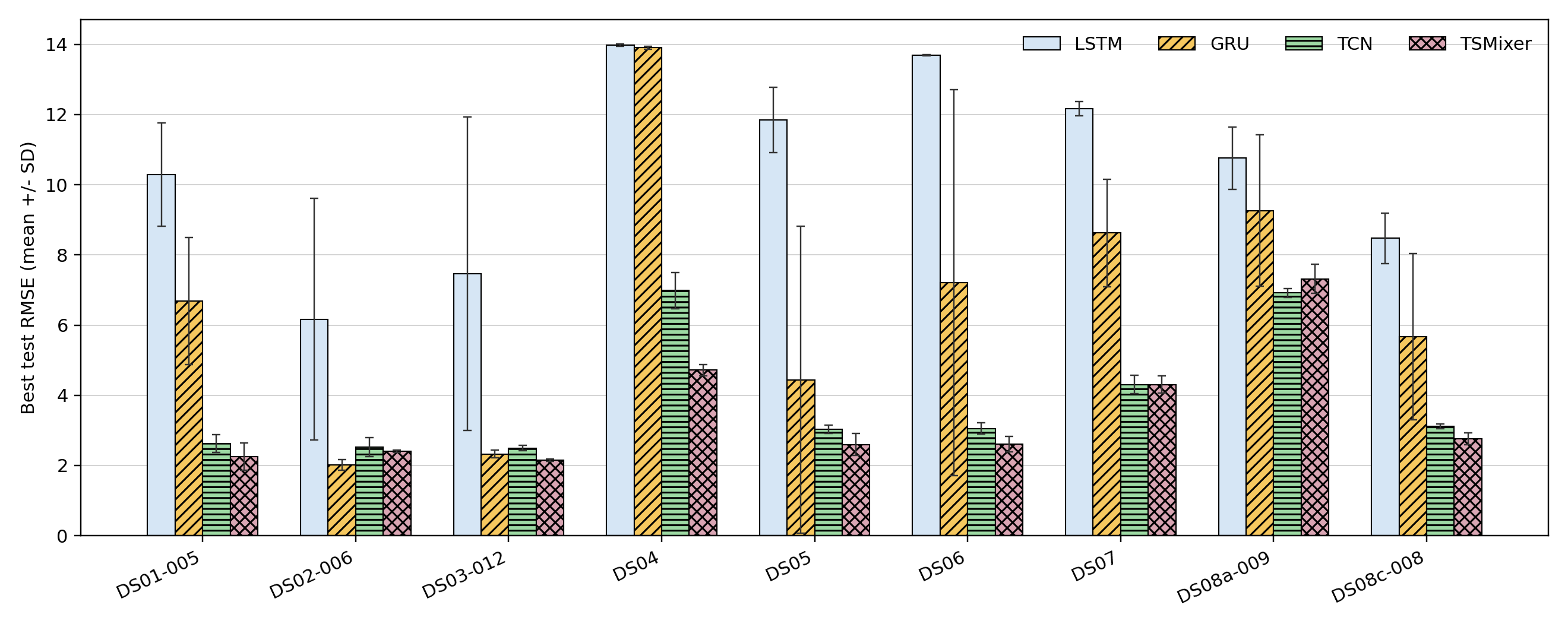}
\caption{Best test RMSE by dataset under the main $\eta=5\%$ setting. Bars show means and error bars show standard deviation across five repeated runs.}
\label{fig:eta5-rmse}
\end{figure}

Figure~\ref{fig:eta5-rmse} highlights DS04 and DS08a-009 as more difficult cases in this benchmark setting. This pattern is consistent with the benchmark design. Once cruise rows are isolated and leakage-prone metadata are removed, dataset-specific degradation and operating-history differences still matter. \datasetname\ preserves this difference rather than make all datasets equally easy.

\subsection{Ablation Studies}\label{sec:ablation-studies}

\subsubsection{Flight-stage sensitivity}
This ablation examines whether model ranking and repeat-to-repeat stability change when the flight-stage constraint is removed. The cruise-stage family uses the released \datasetname\ mask and samples windows only from accepted cruising intervals. The no-mask family keeps the same feature rule, target construction, and model settings, but samples windows from complete flight cycles that may include climb, cruise, descent, and transition segments. Because the two families contain different window populations, the comparison is interpreted as a flight-stage sensitivity analysis rather than a direct leaderboard comparison.

\begin{figure}[H]
\centering
\includegraphics[width=\linewidth]{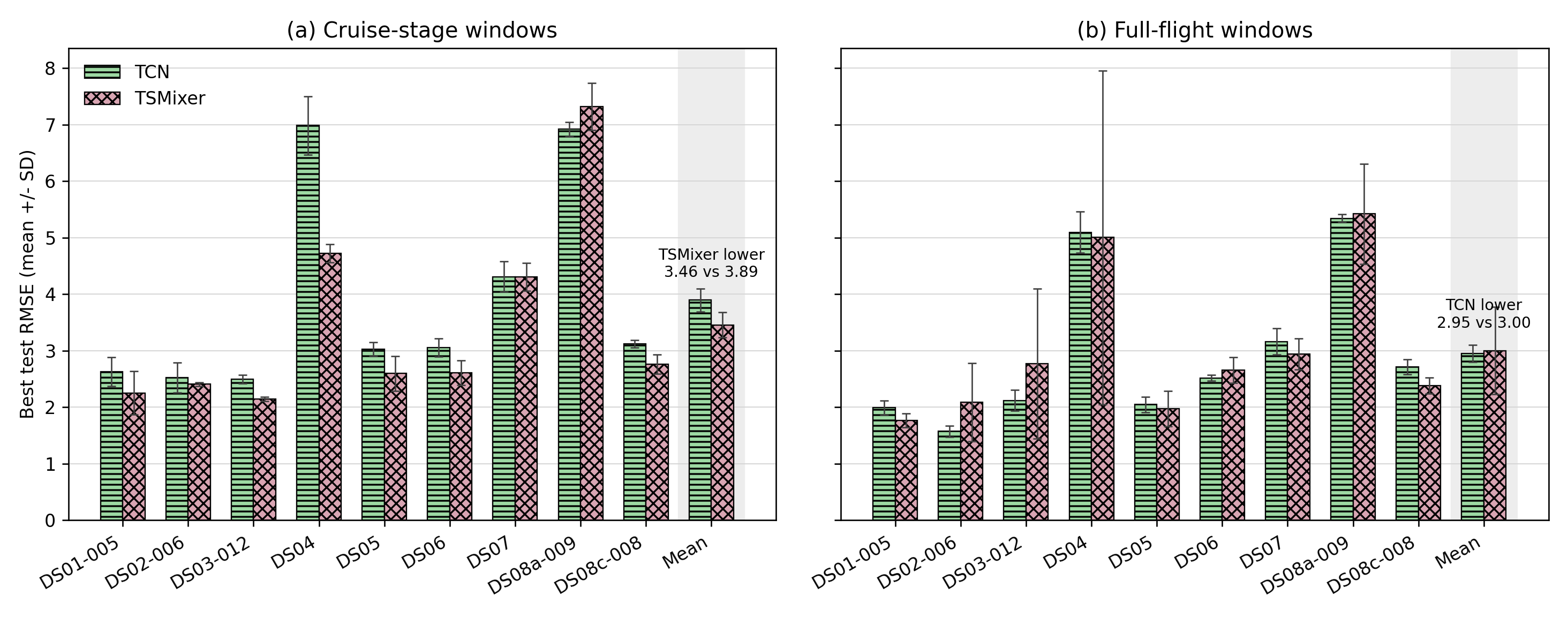}
\caption{TCN and TSMixer ranking under (a) cruise-stage and (b) no-mask full-flight experiment families. Dataset bars show mean best test RMSE with standard deviation across five repeated runs. The shaded Mean group shows the mean of subdataset RMSE means, with the error bar equal to the average repeat-to-repeat standard deviation across subdatasets.}
\label{fig:stage-ranking}
\end{figure}

Figure~\ref{fig:stage-ranking} compares TCN and TSMixer, the two strongest baselines shown in Table~\ref{tab:benchmark-results}. Under the cruise-stage mask, TSMixer has the lower average RMSE, $3.46$ versus $3.89$ for TCN. In the no-mask family, the ordering reverses slightly. TCN averages $2.95$ and TSMixer averages $3.00$.

The ranking reversal indicates that the no-mask setting changes the model comparison by adding operating-regime variation to the prediction windows. Under the full-flight window population, TCN gains more than TSMixer from the added climb, cruise, descent, and transition patterns.
This supports the role of \benchmark\ as a controlled construction because fixing the flight stage separates cruise-stage RUL prediction from performance gains associated with mixed-regime inputs.

Table~\ref{tab:stage-sensitivity-deviation} further shows that the effect is architecture-dependent. TSMixer has a clear increase in repeat-to-repeat deviation when moving from cruise-stage windows to full-flight windows. Its average repeat standard deviation rises from $0.22$ to $0.77$, with the largest increases on DS03-012, DS04, and DS08a-009. TCN does not show the same instability pattern, with repeat standard deviation changing from $0.21$ under cruise-stage windows to $0.16$ under full-flight windows. This suggests that TSMixer is more sensitive to mixed operating regimes under the current setup, and that stronger regime-aware modeling may be needed when it is applied to full-flight inputs.

\input{tables/stage_sensitivity_deviation_table}

The window counts show that this diagnostic changes both flight-stage composition and window population size.
The cruise-stage setting uses $1{,}101{,}166$ train windows and $614{,}293$ test windows, while the no-mask full-flight setting uses $4{,}312{,}736$ train windows and $2{,}490{,}297$ test windows. The assessment therefore shows how model ranking and stability change when the benchmark constraint is removed, rather than a direct result comparison.
It shows that model ranking and training stability can change when flight-stage control is removed, which motivates reporting cruise-stage results separately from full-flight results.

\subsubsection{Window stride versus sample-rate downsampling}
Table~\ref{tab:tsmixer-stride-sample} compares two temporal downscaling methods for TSMixer under the cruise-stage mask. The sample-rate setting downsamples by a factor of 10 with stride 1, while the window-stride setting keeps native-resolution windows with dataset-specific train/test strides. The custom-stride setting uses fewer training windows overall, $353{,}753$ versus $360{,}938$, and more test windows, $178{,}058$ versus $176{,}425$. Even under that slightly less favorable train/test count balance, it improves the average TSMixer RMSE from $3.85$ to $3.69$ and is better on DS03-012, DS04, DS05, DS06, and DS08c-008.

\input{tables/tsmixer_stride_sample_ablation_table}

The result supports window stride as the benchmark's downscaling method. Window stride provides a finer adjustment range than sample-rate downsampling. It can change the number of extracted windows without changing the native temporal spacing inside each 256-sample input. Sample-rate downsampling instead changes the physical time span represented by a fixed-length input and offers a coarser control over usable window counts. The custom-stride setting further shows this flexibility, because train and test splits can use different stride values to meet a target data budget.

\subsubsection{RUL-cap sensitivity}
The $\eta=10\%$ and $\eta=0\%$ settings evaluate sensitivity to the RUL-cap threshold. They all use \datasetname\, feature rule, and windowing rule as the main $\eta=5\%$ setting, but apply different dataset-wise caps for RUL target construction. The resulting RMSE values change because both the target scale and the regression target are modified. These results therefore characterize benchmark-setting sensitivity rather than intrinsic dataset difficulty.

\begin{figure}[H]
\centering
\includegraphics[width=0.76\linewidth]{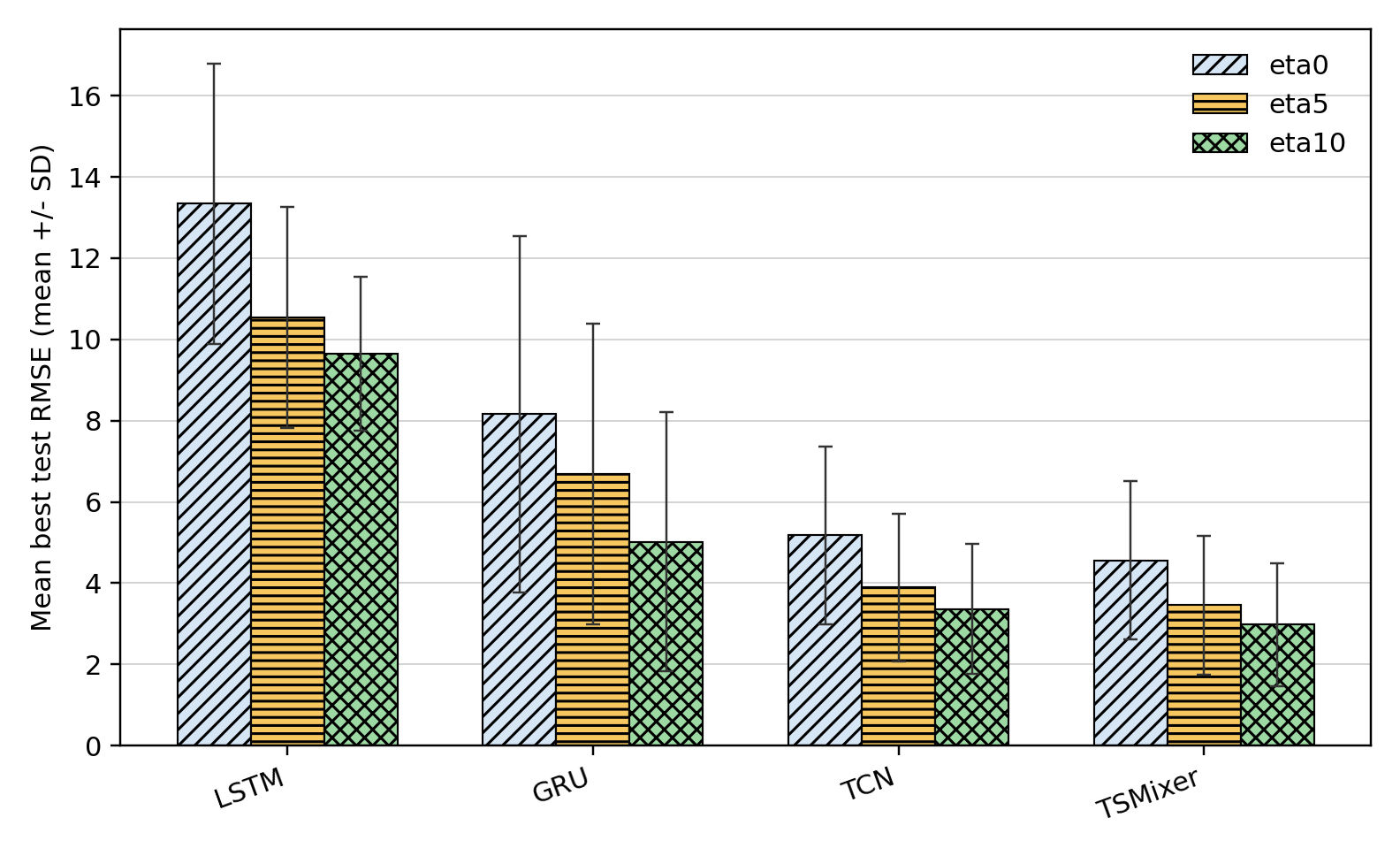}
\caption{Average RMSE sensitivity to the RUL-cap threshold. Bars show means across the nine accessible subdatasets and error bars show standard deviation across dataset-level means.}
\label{fig:rul-cap-sensitivity}
\end{figure}

Table~\ref{tab:benchmark-results} and Figure~\ref{fig:rul-cap-sensitivity} show that cap choice has a visible effect on RMSE and score. The $\eta=10\%$ setting gives smaller absolute RMSE because it uses a stricter cap, while $\eta=0\%$ gives larger values because it uses the full minimum abnormal length as the cap. This supports reporting the cap as part of the setting name. Results obtained with different cap rules should not be merged into a single comparison.

Together, the ablation studies support three findings. First, flight-stage selection affects model ranking and stability, which motivates reporting the cruise-stage setting separately from no-mask full-flight results. Second, window stride provides a practical temporal downscaling choice because it adjusts window density while preserving native-resolution within-window dynamics. Third, the RUL-cap threshold changes reported RMSE and should be treated as part of the benchmark setting.

\subsection{Learning dynamics}
\begin{figure}[H]
\centering
\includegraphics[width=0.76\linewidth]{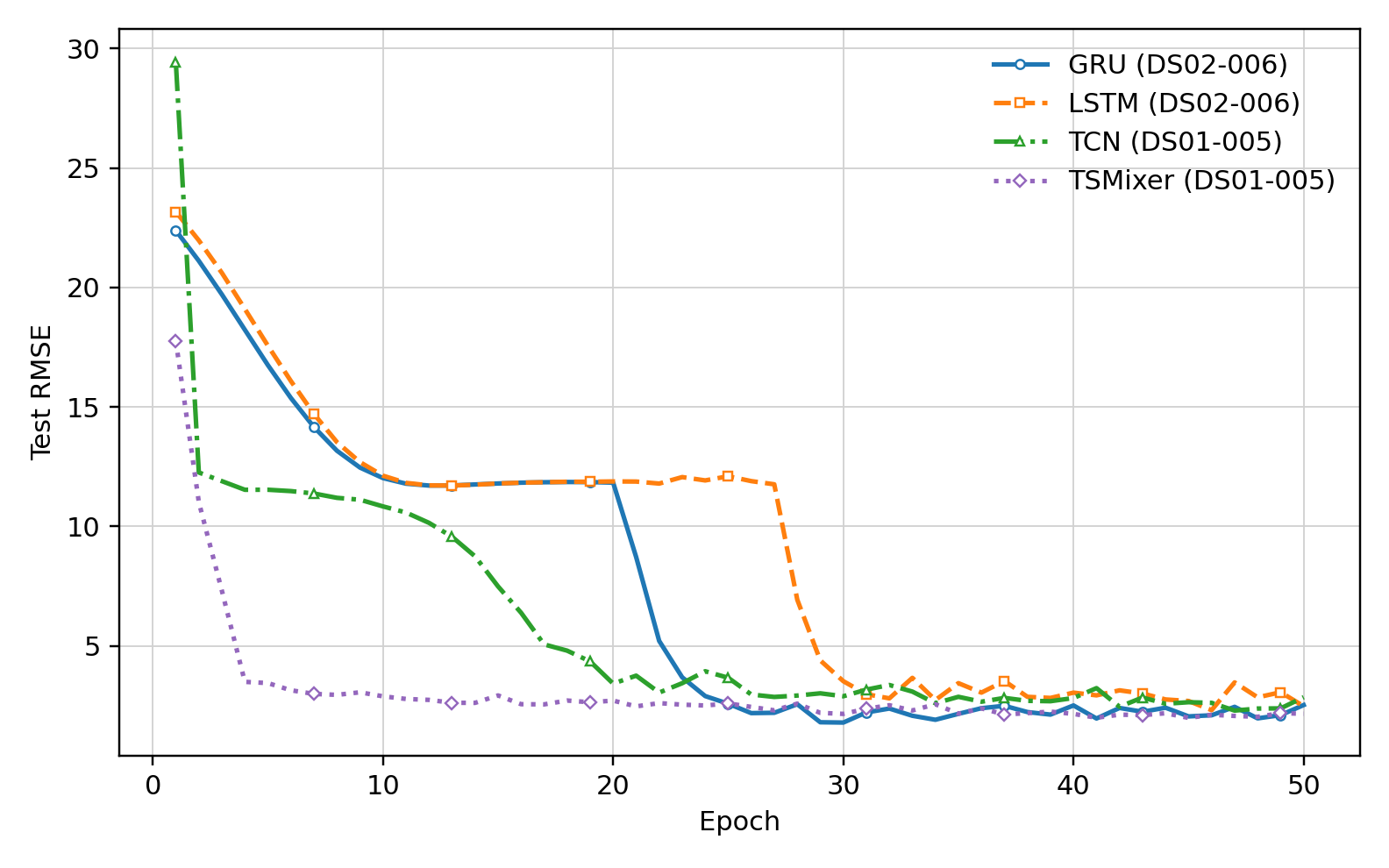}
\caption{Representative $\eta=5\%$ learning curves. The plotted runs show the relation between optimization progress and the best test RMSE used in Table~\ref{tab:benchmark-results}.}
\label{fig:learning-curves}
\end{figure}

The learning curves provide a useful check beyond final numbers. Representative train/test trajectories are shown in Figure~\ref{fig:learning-curves}. The curves show that the reported minima are reached through stable optimization and are not isolated scalar values. They also indicate that simple sequence baselines can fit the cruise-only windows without requiring flight-class or health-state inputs.

%% file: tables/benchmark_results_table.tex
\begingroup
\setlength{\LTcapwidth}{\textwidth}
\setlength{\tabcolsep}{2.4pt}
\renewcommand{\arraystretch}{0.95}
\begin{longtable}{>{\scriptsize}l>{\scriptsize}l>{\scriptsize}c>{\scriptsize}c>{\scriptsize}c>{\scriptsize}c>{\scriptsize}c>{\scriptsize}c}
\caption{Dataset-specific benchmark results under the three RUL-cap settings. RMSE is reported in cycles. Saxena score is reported as $\times 10^4$ and is measured at the epoch with the best test RMSE. Values are mean $\pm$ standard deviation across five repeated runs.}
\label{tab:benchmark-results}\\
\toprule
{\scriptsize Dataset} & {\scriptsize Model} & \multicolumn{2}{c}{\scriptsize $\eta=0\%$} & \multicolumn{2}{c}{\scriptsize $\eta=5\%$} & \multicolumn{2}{c}{\scriptsize $\eta=10\%$} \\
\cmidrule(lr){3-4} \cmidrule(lr){5-6} \cmidrule(lr){7-8}
 & & {\scriptsize RMSE} & {\scriptsize Score} & {\scriptsize RMSE} & {\scriptsize Score} & {\scriptsize RMSE} & {\scriptsize Score} \\
\midrule
\endfirsthead
\caption[]{Dataset-specific benchmark results under the three RUL-cap settings. RMSE is reported in cycles. Saxena score is reported as $\times 10^4$ and is measured at the epoch with the best test RMSE. Values are mean $\pm$ standard deviation across five repeated runs. (continued)}\\
\toprule
{\scriptsize Dataset} & {\scriptsize Model} & \multicolumn{2}{c}{\scriptsize $\eta=0\%$} & \multicolumn{2}{c}{\scriptsize $\eta=5\%$} & \multicolumn{2}{c}{\scriptsize $\eta=10\%$} \\
\cmidrule(lr){3-4} \cmidrule(lr){5-6} \cmidrule(lr){7-8}
 & & {\scriptsize RMSE} & {\scriptsize Score} & {\scriptsize RMSE} & {\scriptsize Score} & {\scriptsize RMSE} & {\scriptsize Score} \\
\midrule
\endhead
\endfoot
\bottomrule
\endlastfoot
DS01-005 & LSTM & $14.61\pm0.00$ & $21.61\pm0.07$ & $10.30\pm1.47$ & $10.29\pm2.85$ & $9.89\pm0.10$ & $10.14\pm0.45$ \\
 & GRU & $6.97\pm2.41$ & $4.63\pm2.10$ & $6.69\pm1.81$ & $4.41\pm1.59$ & $2.37\pm1.15$ & $1.14\pm0.82$ \\
 & TCN & $3.41\pm0.17$ & $1.82\pm0.18$ & $2.63\pm0.25$ & $1.25\pm0.14$ & $2.20\pm0.14$ & $1.01\pm0.10$ \\
 & TSMixer & $2.66\pm0.15$ & $1.27\pm0.10$ & $2.25\pm0.38$ & $1.00\pm0.21$ & $1.76\pm0.06$ & $0.74\pm0.02$ \\
DS02-006 & LSTM & $6.46\pm5.24$ & $3.53\pm3.99$ & $6.17\pm3.45$ & $2.89\pm2.23$ & $7.20\pm2.74$ & $3.36\pm1.86$ \\
 & GRU & $2.83\pm0.25$ & $0.77\pm0.11$ & $2.02\pm0.15$ & $0.51\pm0.03$ & $1.86\pm0.52$ & $0.46\pm0.15$ \\
 & TCN & $3.29\pm0.24$ & $1.01\pm0.10$ & $2.52\pm0.26$ & $0.73\pm0.10$ & $2.19\pm0.21$ & $0.60\pm0.07$ \\
 & TSMixer & $3.17\pm0.14$ & $0.98\pm0.08$ & $2.41\pm0.03$ & $0.65\pm0.02$ & $2.24\pm0.16$ & $0.60\pm0.05$ \\
DS03-012 & LSTM & $12.59\pm1.96$ & $23.58\pm6.89$ & $7.46\pm4.47$ & $11.58\pm8.88$ & $7.27\pm3.01$ & $9.72\pm5.51$ \\
 & GRU & $3.39\pm0.50$ & $2.72\pm0.54$ & $2.33\pm0.11$ & $1.59\pm0.08$ & $3.08\pm1.73$ & $2.42\pm2.03$ \\
 & TCN & $3.42\pm0.11$ & $2.89\pm0.16$ & $2.50\pm0.07$ & $1.93\pm0.06$ & $2.28\pm0.09$ & $1.68\pm0.11$ \\
 & TSMixer & $2.99\pm0.09$ & $2.33\pm0.09$ & $2.14\pm0.03$ & $1.54\pm0.04$ & $1.94\pm0.11$ & $1.39\pm0.09$ \\
DS04 & LSTM & $17.33\pm0.19$ & $36.78\pm0.76$ & $13.97\pm0.03$ & $24.49\pm0.65$ & $11.92\pm0.22$ & $18.14\pm0.76$ \\
 & GRU & $17.28\pm0.10$ & $37.54\pm1.15$ & $13.90\pm0.04$ & $24.54\pm0.52$ & $11.94\pm0.08$ & $18.21\pm0.46$ \\
 & TCN & $9.28\pm0.42$ & $30.50\pm37.71$ & $6.98\pm0.52$ & $6.85\pm1.09$ & $5.99\pm0.63$ & $5.47\pm1.17$ \\
 & TSMixer & $6.90\pm0.31$ & $7.63\pm2.56$ & $4.72\pm0.16$ & $3.51\pm0.28$ & $4.25\pm0.22$ & $2.97\pm0.26$ \\
DS05 & LSTM & $14.93\pm0.92$ & $20.06\pm3.11$ & $11.85\pm0.94$ & $13.25\pm1.66$ & $10.67\pm0.07$ & $10.93\pm0.08$ \\
 & GRU & $8.02\pm3.49$ & $6.13\pm3.92$ & $4.44\pm4.38$ & $3.64\pm5.54$ & $2.89\pm0.74$ & $1.44\pm0.54$ \\
 & TCN & $4.34\pm0.16$ & $2.47\pm0.17$ & $3.03\pm0.12$ & $1.49\pm0.08$ & $2.33\pm0.14$ & $1.07\pm0.09$ \\
 & TSMixer & $3.85\pm0.32$ & $2.05\pm0.20$ & $2.60\pm0.31$ & $1.23\pm0.18$ & $2.19\pm0.14$ & $1.05\pm0.11$ \\
DS06 & LSTM & $16.69\pm0.67$ & $25.15\pm2.78$ & $13.69\pm0.01$ & $17.13\pm0.18$ & $12.11\pm0.02$ & $13.84\pm0.24$ \\
 & GRU & $11.33\pm5.75$ & $13.11\pm10.17$ & $7.21\pm5.51$ & $7.05\pm7.43$ & $4.86\pm4.36$ & $3.98\pm5.50$ \\
 & TCN & $4.76\pm0.22$ & $2.80\pm0.12$ & $3.05\pm0.16$ & $1.53\pm0.10$ & $2.59\pm0.24$ & $1.21\pm0.12$ \\
 & TSMixer & $4.29\pm0.37$ & $2.21\pm0.15$ & $2.61\pm0.22$ & $1.24\pm0.14$ & $2.15\pm0.22$ & $0.99\pm0.17$ \\
DS07 & LSTM & $15.54\pm0.04$ & $23.88\pm0.28$ & $12.17\pm0.20$ & $15.14\pm0.69$ & $10.59\pm0.15$ & $11.84\pm0.57$ \\
 & GRU & $8.53\pm2.74$ & $7.01\pm4.22$ & $8.63\pm1.53$ & $6.93\pm2.45$ & $6.54\pm2.42$ & $5.05\pm3.65$ \\
 & TCN & $6.39\pm0.52$ & $4.09\pm0.58$ & $4.31\pm0.27$ & $2.32\pm0.20$ & $4.02\pm0.41$ & $1.97\pm0.26$ \\
 & TSMixer & $5.50\pm0.47$ & $3.13\pm0.25$ & $4.30\pm0.25$ & $2.19\pm0.33$ & $3.42\pm0.08$ & $1.66\pm0.08$ \\
DS08a-009 & LSTM & $11.37\pm2.11$ & $19.79\pm5.53$ & $10.76\pm0.89$ & $16.61\pm1.98$ & $9.54\pm0.41$ & $13.05\pm1.28$ \\
 & GRU & $9.41\pm2.36$ & $14.89\pm5.77$ & $9.26\pm2.15$ & $13.38\pm4.18$ & $7.32\pm1.05$ & $8.92\pm1.99$ \\
 & TCN & $7.92\pm0.25$ & $12.78\pm1.17$ & $6.92\pm0.13$ & $9.00\pm0.38$ & $6.02\pm0.05$ & $6.71\pm0.13$ \\
 & TSMixer & $8.29\pm0.38$ & $13.60\pm2.13$ & $7.32\pm0.42$ & $10.12\pm1.27$ & $6.43\pm0.28$ & $7.61\pm0.59$ \\
DS08c-008 & LSTM & $10.56\pm1.02$ & $7.66\pm1.33$ & $8.47\pm0.73$ & $5.27\pm0.86$ & $7.70\pm0.39$ & $4.64\pm0.36$ \\
 & GRU & $5.68\pm3.34$ & $3.19\pm2.90$ & $5.68\pm2.37$ & $2.99\pm1.91$ & $4.25\pm2.62$ & $2.06\pm1.86$ \\
 & TCN & $3.85\pm0.15$ & $1.61\pm0.09$ & $3.12\pm0.06$ & $1.21\pm0.03$ & $2.66\pm0.11$ & $0.98\pm0.06$ \\
 & TSMixer & $3.37\pm0.13$ & $1.33\pm0.07$ & $2.76\pm0.17$ & $0.98\pm0.09$ & $2.43\pm0.07$ & $0.81\pm0.02$ \\
\end{longtable}
\endgroup

%% file: tables/stage_sensitivity_deviation_table.tex
\begin{table}[H]
\centering
\caption{Subdataset-specific repeat-to-repeat RMSE deviation for TCN and TSMixer under cruise-stage and no-mask full-flight windows. Values are standard deviations across five repeated runs. $\Delta$SD is full-flight minus cruise-stage repeat SD for the same model and subdataset.}
\label{tab:stage-sensitivity-deviation}
\scriptsize
\resizebox{\linewidth}{!}{%
\begin{tabular}{lcccccc}
\toprule
Dataset & TCN cruise SD & TCN full SD & TCN $\Delta$SD & TSMixer cruise SD & TSMixer full SD & TSMixer $\Delta$SD \\
\midrule
DS01-005 & 0.25 & 0.13 & $-0.13$ & 0.38 & 0.12 & $-0.26$ \\
DS02-006 & 0.26 & 0.10 & $-0.17$ & 0.03 & 0.69 & $+0.66$ \\
DS03-012 & 0.07 & 0.19 & $+0.11$ & 0.03 & 1.33 & $+1.30$ \\
DS04 & 0.52 & 0.37 & $-0.15$ & 0.16 & 2.95 & $+2.79$ \\
DS05 & 0.12 & 0.14 & $+0.01$ & 0.31 & 0.31 & $+0.00$ \\
DS06 & 0.16 & 0.05 & $-0.11$ & 0.22 & 0.23 & $+0.00$ \\
DS07 & 0.27 & 0.23 & $-0.04$ & 0.25 & 0.28 & $+0.03$ \\
DS08a-009 & 0.13 & 0.07 & $-0.05$ & 0.42 & 0.89 & $+0.47$ \\
DS08c-008 & 0.06 & 0.13 & $+0.07$ & 0.17 & 0.14 & $-0.03$ \\
Mean & 0.21 & 0.16 & $-0.05$ & 0.22 & 0.77 & $+0.55$ \\
\bottomrule
\end{tabular}
}
\end{table}

%% file: tables/tsmixer_stride_sample_ablation_table.tex
\begin{table}[H]
\centering
\caption{TSMixer comparison between sample-rate downsampling and native-resolution custom stride. Window counts are train/test. RMSE values are mean $\pm$ standard deviation across five repeated runs. The final row reports total windows and mean $\pm$ standard deviation across dataset-level RMSE means. Negative $\Delta$ favors custom stride.}
\label{tab:tsmixer-stride-sample}
\scriptsize
\resizebox{\linewidth}{!}{%
\begin{tabular}{lcccccc}
\toprule
Dataset & Sample10 windows & Sample10 RMSE & Custom stride & Custom windows & Custom RMSE & $\Delta$ \\
\midrule
DS01-005 & 33,959/16,106 & $2.03\pm0.29$ & 35/40 & 33,496/16,124 & $2.16\pm0.16$ & $+0.13$ \\
DS02-006 & 76,557/9,416 & $2.94\pm0.14$ & 23/38 & 74,094/9,535 & $3.23\pm0.15$ & $+0.29$ \\
DS03-012 & 32,674/33,901 & $2.75\pm0.14$ & 40/30 & 32,141/33,951 & $2.20\pm0.14$ & $-0.55$ \\
DS04 & 53,354/22,529 & $9.54\pm2.24$ & 29/36 & 51,597/22,667 & $8.02\pm1.61$ & $-1.52$ \\
DS05 & 31,204/16,870 & $2.73\pm0.23$ & 35/36 & 30,946/17,254 & $2.48\pm0.34$ & $-0.25$ \\
DS06 & 30,981/16,870 & $2.73\pm0.32$ & 35/36 & 30,275/17,103 & $2.59\pm0.47$ & $-0.14$ \\
DS07 & 32,090/19,277 & $2.84\pm0.37$ & 34/35 & 31,529/19,533 & $3.21\pm0.25$ & $+0.37$ \\
DS08a-009 & 21,343/33,379 & $6.14\pm0.43$ & 51/28 & 21,131/33,785 & $6.63\pm0.32$ & $+0.49$ \\
DS08c-008 & 48,776/8,077 & $2.94\pm0.41$ & 23/58 & 48,544/8,106 & $2.69\pm0.25$ & $-0.25$ \\
Mean/total & 360,938/176,425 & $3.85\pm2.43$ & -- & 353,753/178,058 & $3.69\pm2.12$ & $-0.16$ \\
\bottomrule
\end{tabular}
}
\end{table}

%% file: sections/07_discussion.tex
\benchmark\ isolates a stable flight stage and therefore reduces flight-stage composition as a confounding factor in model comparison. The results shown in Table~\ref{tab:benchmark-results} support this framing. Several ordinary sequence encoders reach low RMSE without using health state, flight class, unit id, cycle id, or sample index as predictive features. This makes the benchmark useful as a controlled sub-benchmark, but it also narrows the operational question. Climb and descent may contain useful degradation information that is intentionally removed here. Full-flight N-CMAPSS evaluation remains necessary for models intended to operate across the entire flight envelope.

The RUL-cap policy is also a setting choice rather than a directly observed detectability threshold. The exponential abnormal-degradation law motivates a smaller cap than the full abnormal period, but $\eta$ should be treated as a benchmark setting. The $\eta=0\%$, $\eta=5\%$, and $\eta=10\%$ results differ enough that they should not be combined into one leaderboard. Reporting each $\eta$ setting separately makes this choice clear.

Finally, the cruise mask is fixed given the source HDF5 files and method settings. This supports reproducibility, but the extraction method still contains assumptions. Cruise is represented as a long high-altitude common-level plateau, and the final accepted interval is selected by duration and altitude stability. These assumptions are appropriate for a cruise-stage benchmark and should not be generalized to every flight stage.

%% file: sections/08_conclusion.tex
This paper presented \benchmark, a cruise-stage RUL benchmark derived from N-CMAPSS, and \datasetname, a cruise-stage mask artifact that stores cycle-local cruising intervals for the nine accessible subdatasets. The benchmark applies a fixed protocol to masked cruise-stage rows. Scenario descriptors and measured sensors are used as inputs, virtual sensors, health parameters, and auxiliary metadata are excluded from the feature tensor, windows preserve the native time resolution, and RUL labels use dataset-wise caps. Baseline experiments with LSTM, GRU, TCN, and TSMixer show that the cruise-stage setting retains learnable RUL structure, with TSMixer obtaining the lowest average RMSE and Saxena score under \mainsetting.

The ablation studies further show that flight-stage selection, temporal downscaling method, and RUL-cap threshold affect reported results and should be treated as explicit benchmark settings. With its fixed cruise-stage protocol, \benchmark\ supports controlled RUL model comparison. \datasetname\ also provides a stage-specific data foundation for future transfer-learning and domain-adaptation studies. Future work can extend the same mask-based design to additional operating regimes, evaluate cross-stage transfer, and test specialized prognostics architectures under fixed data and target-construction rules.

%% file: sections/09_reproducibility.tex
The open-source package provides the released cruise mask and all tools needed to regenerate and inspect it. The raw N-CMAPSS dataset files are not redistributed. Users place the official HDF5 files in the expected local data directory, then either use the released mask directly or regenerate it with the provided exporter.

The released mask file is:
\begin{quote}
\texttt{dataset/cruise\_mask/common\_altitude\_cruise\_mask.h5}
\end{quote}
It contains 7473 cycle rows and 7053 accepted cruise cycles. The relevant HDF5 attributes are:
\begin{center}
\begin{tabular}{ll}
\toprule
Attribute & Value \\
\midrule
profile & \texttt{common\_altitude} \\
method & \texttt{common\_altitude} \\
created at & \texttt{2026-04-27T18:44:21} \\
interval unit & \texttt{cycle\_local\_sample\_index} \\
interval end & \texttt{inclusive} \\
\bottomrule
\end{tabular}
\end{center}

For each experiment, reports should include the source dataset list, cruise-mask settings, window size, downscaling method, feature set, dataset-wise RUL caps and model hyperparameters.